\documentclass{article}
\usepackage{spconf,amsmath,graphicx,hyperref}
\usepackage{multirow}
\usepackage{booktabs}
\usepackage{colortbl}
\usepackage{xcolor} 
\usepackage{nccmath} 
\usepackage{algorithmic}
\usepackage{algorithm}
\usepackage{amssymb}
\usepackage{setspace}

\title{GUI-ARP: Enhancing Grounding with Adaptive Region Perception \\for GUI Agents}
\name{
\parbox{\textwidth}{\centering
    Xianhang Ye$^{\ast,1,3}$, Yiqing Li$^{\ast,2,3}$, 
    Wei Dai$^{3}$, 
    Miancan Liu$^{2,3}$, 
    Ziyuan Chen$^{3,4}$, \\
    \textit{Zhangye Han$^{3,5}$},
    \textit{Hongbo Min$^{\ddag,3}$}, 
    \textit{Jinkui Ren$^{3}$}, 
    \textit{Xiantao Zhang$^{3}$}, 
    \textit{Wen Yang$^{\dag,1}$},
    \textit{Zhi Jin$^{\dag,2}$}
    \thanks{* Equal contribution. Work done during an internship at Alibaba Group.}
    \thanks{$\dag$ Corresponding author. E-mail: \emph{yangwen@whu.edu.cn.}}
    \thanks{$\ddag$ Project Leader.}
    }
}

\address{
  \textsuperscript{1}Wuhan University\quad
  \textsuperscript{2}Sun Yat-sen University\quad
  \textsuperscript{3}Alibaba Group \\
  \hspace{\parindent}  
  \textsuperscript{4}East China Normal University\quad
  \textsuperscript{5}University of Electronic Science and Technology of China
}
\begin{document}
\maketitle
\begin{abstract}
Existing GUI grounding methods often struggle with fine-grained localization in high-resolution screenshots. To address this, we propose GUI-ARP, a novel framework that enables adaptive multi-stage inference. Equipped with the proposed Adaptive Region Perception (ARP) and Adaptive Stage Controlling (ASC), GUI-ARP dynamically exploits visual attention for cropping task‑relevant regions and adapts its inference strategy, performing a single-stage inference for simple cases and a multi-stage analysis for more complex scenarios. This is achieved through a two-phase training pipeline that integrates supervised fine-tuning with reinforcement fine-tuning based on Group Relative Policy Optimization (GRPO). Extensive experiments demonstrate that the proposed GUI-ARP achieves state-of-the-art performance on challenging GUI grounding benchmarks, with a 7B model reaching $60.8\%$ accuracy on ScreenSpot-Pro and $30.9\%$ on UI-Vision benchmark. Notably, GUI-ARP-7B demonstrates strong competitiveness against open-source 72B models (UI-TARS-72B at $38.1\%$) and proprietary models.
\end{abstract}
\maketitle
\section{Introduction}
\label{sec:intro}
Graphical User Interface (GUI) agents \cite{abuelsaad2024agent-e,agent-s,lin2025showui,qin2025ui-tars} have become a prominent research topic due to their ability to automate complex tasks, such as sending emails and booking hotels, based on natural language instructions. A key technical challenge in this domain is GUI grounding, which requires the accurate localization of actionable elements within a user interface guided by natural language instructions. 

\begin{figure}[t]
\begin{minipage}[b]{1.0\linewidth}
  \centering
  \centerline{\includegraphics[width=8.5cm]{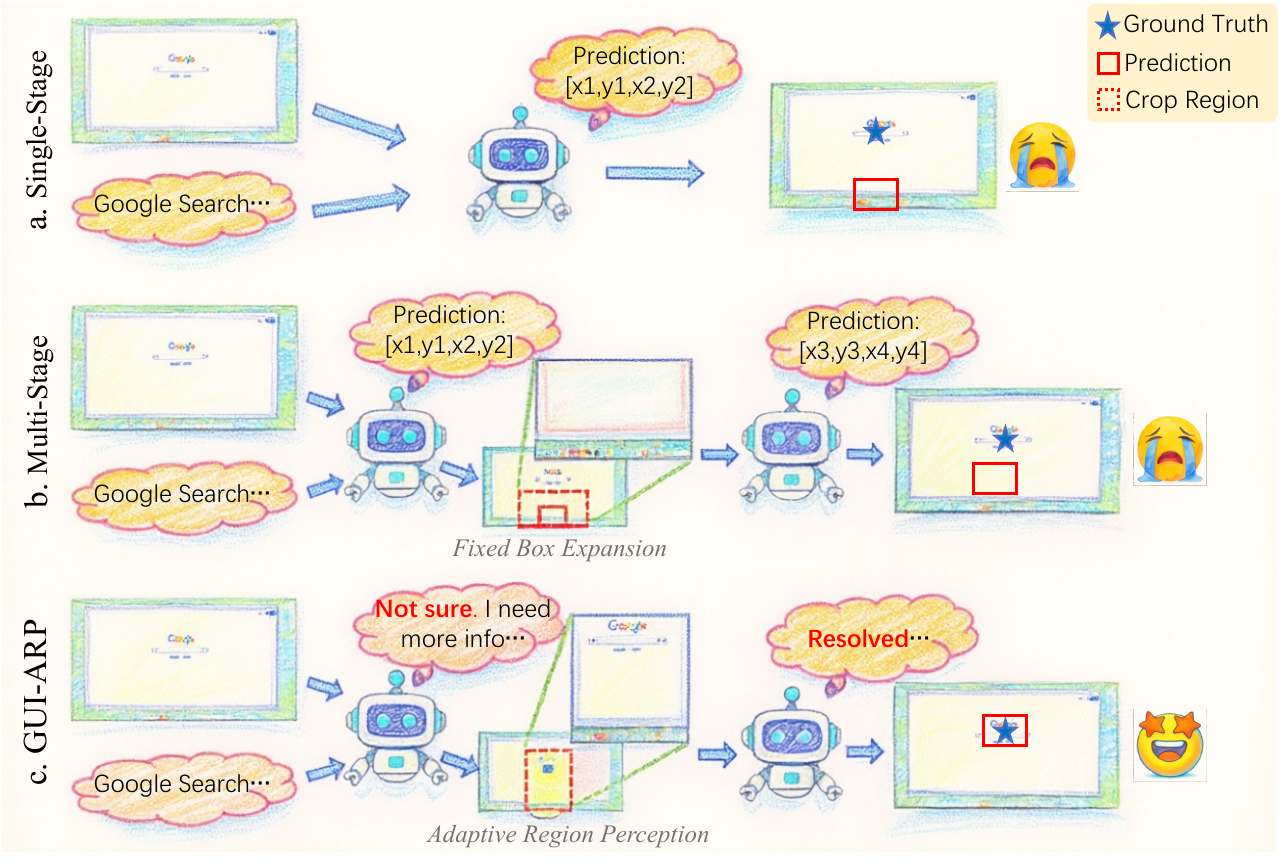}}
  \vspace{-0.7cm}
\end{minipage}
\caption{\small{Comparison of grounding strategies. (a) Single‑stage: Efficient but struggles in complex cases. (b) Multi‑stage: Finer localization but relies on heuristic box expansion. (c) GUI‑ARP (ours): Adaptively controls inference stages and leverages visual attention for precise region perception.}}
\label{fig:res}
\end{figure}

Most existing GUI grounding methods \cite{lu2025ui-r1,liu2025infigui-g1,luo2025gui-r1,tang2025gui-g2} are built upon Vision-Language Models (VLMs) and employ Reinforcement Fine-Tuning (RFT) frameworks to perform text-based coordinate generation. Conversely, GUI-Actor \cite{wu2025guiactor} generates attention maps through large-scale Supervised Fine-Tuning (SFT) as a substitute for text coordinates. While these methods perform well in many scenarios, their element localization is suboptimal in complex situations, particularly with high-resolution user interfaces. This limitation stems from the patchify operation in current ViT \cite{dosovitskiy2021ViT}, which compromises fine-grained spatial perception and makes accurate single-stage reasoning challenging (Fig.~\ref{fig:res} a).

Inspired by the human visual localization habit of ``glance-and-focus" in complex scenes, we introduce a novel multi-stage grounding paradigm to advance from passive perception to active visual cognition. 
Some studies \cite{liu2024chain-of-spot, zhang2025chain-of-focus, shen2024zoomeye} have attempted to incorporate this human visual behavior into VLMs to accomplish visual question answering tasks. In the GUI grounding domain, the recent work R-VLM \cite{park2025r-vlm} also applies the multi-stage perception. However, these paradigms rely on manually designed zoom-in strategies, such as expanding the predicted bounding box by a fixed factor, which fails to take advantage of the visual perception of the VLMs (Fig.~\ref{fig:res} b). Besides, this multi-stage approach introduces unnecessary overhead and may degrade performance in simple scenarios, as the zoomed-in region lacks sufficient visual cues, compromising the global perception of the model.

To address the aforementioned challenges, we propose a novel grounding paradigm named \textbf{GUI-ARP}, which enables adaptive local perception and stage-aware control. Specifically, instead of formulating grounding as a text generation task, we adapt the attention-based framework \cite{wu2025guiactor}, which provides more transparent insight into the internal visual focus of the model. We further propose Adaptive Region Perception (ARP) and Adaptive Stage Controlling (ASC) strategies to facilitate precise region perception and dynamically adjust inference stage based on task complexity. Moreover, we introduce a high-quality dataset and a two-phase training pipeline which combines SFT for cold start and RFT using Group Relative Policy Optimization (GRPO) \cite{shao2024grpo}. Our model intelligently performs single-stage inference for simple cases while executing a coarse-to-fine multi-stage for challenging ones. 

Our contributions can be summarized as follows: 
\begin{itemize}
    \item We propose GUI-ARP, a novel GUI grounding framework that utilizes adaptive region perception and adaptive stage controlling, advancing the paradigm from passive perception toward active visual cognition.
    \item We collect a high-quality dataset and introduce a two-phase training pipeline including SFT and GRPO, enabling the model to perform adaptive multi-stage inference based on task complexity.
    \item Extensive experiments on public benchmarks demonstrate that our method achieves state-of-the-art performance among 7B models and remains competitive against larger 72B and closed-source models.
\end{itemize}

\section{Method}
\label{sec:method}
\begin{figure*}[t]
\begin{minipage}[b]{1.0\textwidth}
  \centering
  \centerline{\includegraphics[width=0.95\textwidth]{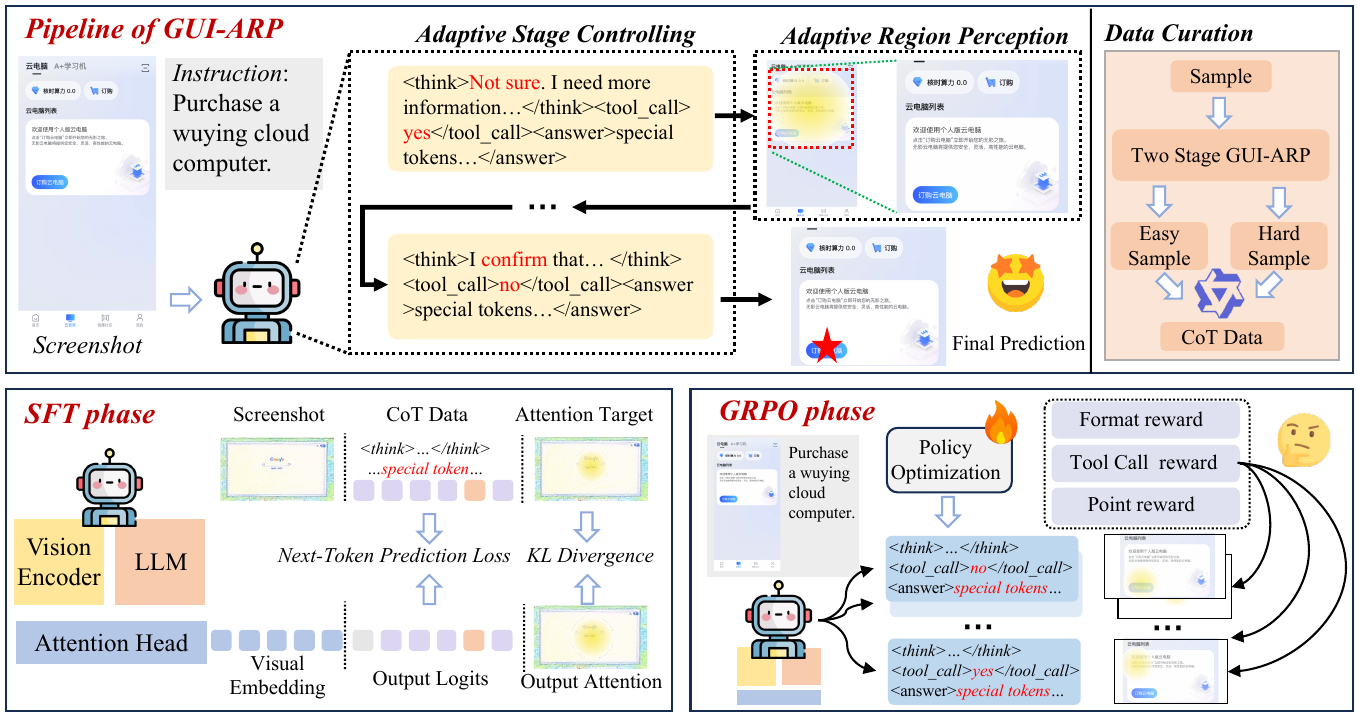}}
\end{minipage}
\vspace{-0.5cm}
\caption{\small{Framework of GUI‑ARP. Given a screenshot and instruction, ASC assesses task difficulty and controls the multi‑stage process. For challenging cases, ARP leverages model attention to extract the most relevant foreground region for further observation. Training proceeds in two stages: SFT with next‑token prediction and KL divergence losses, followed by GRPO with format, tool call, and point rewards.}}
\label{fig:main}
\end{figure*}
An overview of GUI-ARP is shown in Fig.~\ref{fig:main}. In the following sections, we introduce our adaptive region perception, adaptive stage controlling and data curation.
\subsection{Preliminary}
GUI-Actor~\cite{wu2025guiactor} adapts a coordinate-free grounding paradigm by introducing a special contextual anchor token \texttt{<ACTOR>}.  
Given a GUI screenshot and a natural language instruction, the model jointly encodes both modalities, producing hidden representations for all tokens.  
An attention head is then attached to the model, leveraging the final-layer embedding of \texttt{<ACTOR>} as a query to attend over all visual patch tokens.  
The resulting attention weights indicate the spatial relevance of each patch to the target element, thereby directly localizing the actionable region on the interface.  
Our approach extends this paradigm with adaptive region perception and stage-aware control.

\subsection{Adaptive Region Perception}
\label{subsec:arp}
Existing multi‑stage methods often rely on heuristic strategies. For example, R‑VLM~\cite{park2025r-vlm} enlarges the first‑stage predicted bounding box by a fixed ratio, which cannot adapt to varying localization errors. Large deviations from the ground truth require broader crops to capture the target, whereas small deviations benefit from finer crops to avoid redundant background. To this end, we propose an attention‑based Adaptive Region Perception (ARP) approach. In our GUI‑ARP, attention weights encode the spatial relevance of each visual patch to the target. By analyzing the distribution of attention weights, ARP adaptively selects the cropping region, enabling more precise region perception. The process is formulated as follows:
\begin{algorithm}[!h]
    \caption{Adaptive Region Perception (ARP)}
    \label{alg:ARP}
    \renewcommand{\algorithmicrequire}{\textbf{Input:}}
    \renewcommand{\algorithmicensure}{\textbf{Output:}}
    \begin{algorithmic}[1]
        \REQUIRE Attention map $\mathbf{A} \in \mathbb{R}^{H \times W}$, threshold $\tau$, top‑$k$
        \ENSURE Cropping region $\mathcal{R}$
        
        \STATE $\mathcal{S} \gets \{(i,j) \mid \mathbf{A}_{i,j} \ge \tau \cdot \max(\mathbf{A}) \}$ 
        \STATE $\{\mathcal{G}_1, \mathcal{G}_2, \dots\}$ $\gets$ ConnectedComponents($\mathcal{S}$)\\
        \# Split into connected regions
        \STATE $\mathcal{T} \gets$ select top‑$k$ regions with highest scores
        \STATE $P \gets \{\text{GetWeightedCenter}(\mathcal{G}) \mid \mathcal{G} \in \mathcal{T}\}$ 
        \STATE $x_{\min}, y_{\min}, x_{\max}, y_{\max} \gets$ min/max coords in $P$
        \STATE $\mathcal{R} \gets \mathrm{BBox}(x_{\min}, y_{\min}, x_{\max}, y_{\max})$ 
        \RETURN $\mathcal{R}$
    \end{algorithmic}
\end{algorithm}

\subsection{Adaptive Stage Controlling}
While ARP can provide more interpretable candidate target regions, the model lacks the capability to determine whether further observation is necessary. To address this limitation, we propose Adaptive Stage Controlling (ASC), a mechanism that dynamically controls the multi-stage inference process. This approach achieves high precision while significantly improving overall efficiency.

\noindent \textbf{SFT:} We incorporate a Chain-of-Thought (CoT) reasoning process and structured control tokens \texttt{<tool$\_$call>} yes/no \texttt{</tool$\_$call>} to indicate whether ARP should be invoked for finer-grained observation of candidate visual regions. Following \cite{wu2025guiactor}, the overall loss function consists of the Next-Token Prediction (NTP) loss and the attention loss.
\begin{equation}
\begin{aligned}
& \mathcal{L}_{NTP}(\theta)=-\frac{1}{T} \sum_{t=1}^{T} \log P\left(x_{t} \mid x_{0}, x_{1}, \ldots, x_{t-1} ; \theta\right), \\
& \mathcal{L}_{Attn} = \sum_{i=1}^{M} p_i \log \frac{p_i}{a_i}, \quad p_i = \frac{y_i}{\sum_{j=1}^{M} y_j + \epsilon}, \; i = 1, \ldots, M, \\
& \mathcal{L} = \mathcal{L}_{NTP} + \mathcal{L}_{Attn}.
\end{aligned}
\end{equation}
In the NTP loss, $x_t$ denotes the $t$-th token in the output sequence. For the attention supervision, the ground truth attention patch $y_i$ is constructed following \cite{wu2025guiactor}. We denote $p_i$ as the normalized version of $y_i$, $a_i$ as the predicted attention value at the same patch, and $M$ as the total number of patches. The divergence between the predicted attention distribution and the ground truth is measured using the KL divergence.

\noindent \textbf{GRPO:} In the RFT phase, we utilize the GRPO \cite{shao2024grpo} algorithm for policy optimization. Inspired by \cite{tang2025gui-g2}, we introduce Gaussian point rewards $\mathcal{R}_{point}$ to mitigate the sparsity of the rewards. We design the following rule-based rewards to guide the optimization of the policy model.
\begin{equation}
\begin{aligned}
    & \mathcal{R}_{point}=\exp\left(-\frac{1}{2}\left[\frac{(c_x^p-c_x^{gt})^2}{\sigma_x^{gt^2}}+\frac{(c_y^p-c_y^{gt})^2}{\sigma_y^{gt^2}}\right]\right), \\
    & \mathcal{R}_{tool} = 
    \begin{cases}
    1, & \text{if}\quad (\text{tool call = no} \land (c_x^p, c_y^p) \in gt_{bbox}) \\
    & \quad \lor\ (\text{tool call = yes} \land (c_x^p, c_y^p) \notin gt_{bbox}), \\
    0, & \text{otherwise.}
    \end{cases}
    \\
    & \mathcal{R} = \mathcal{R}_{format} + \mathcal{R}_{tool} + \mathcal{R}_{point}.
\end{aligned}
\label{eq:grpo_rewards}
\end{equation}

\noindent where $\mathcal{R}_{format}=1$ if the response adheres to the predefined format and $0$ otherwise. $(c_x^p, c_y^p)$ is predicted point, $(c_x^{gt}, c_y^{gt})$ is the center of the ground truth bounding box $gt_{bbox} = (x_1,y_1,x_2,y_2)$, $\sigma_x^{gt} = \alpha \cdot (x_2 - x_1), \sigma_y^{gt} = \alpha \cdot (y_2 - y_1)$, $\alpha$ is a scaling factor that controls the relative influence of element size on the standard
deviations. $\mathcal{R}_{tool}$ is designed to encourage the model to perform multi-stage grounding when further observation is needed for challenging samples, while avoiding unnecessary refinement on simple cases.

\begin{table*}[t!]
  \centering
  \caption{Performance comparison on ScreenSpot-v1, v2, Pro and UI-Vision. Bold highlights the best results, ``-'' indicates missing values due to unavailable results in the original paper, unreleased model checkpoints, and inference code.}
    \resizebox{\textwidth}{!}{
    \begin{tabular}{lccccccccccccc}
    \toprule
    \multicolumn{1}{c}{\multirow{2}[4]{*}{\textbf{Model}}} & \multicolumn{7}{c}{\textbf{ScreenSpot Pro Accuracy (\%)}} & \multirow{2}[4]{*}{\textbf{SS-v1 Avg.}} & \multirow{2}[4]{*}{\textbf{SS-v2 Avg.}} & \multicolumn{4}{c}{\textbf{UI-Vision Accuracy (\%)}} \\
\cmidrule{2-8} \cmidrule{11-14}  & \textbf{Dev} & \textbf{Creative} & \textbf{CAD} & \textbf{Scientific} & \textbf{Office} & \textbf{OS} & \textbf{Avg.} &    &   &  \textbf{Basic} & \textbf{Fun.} & \textbf{Spatial} & \textbf{Avg.}\\
    \midrule
    \multicolumn{14}{l}{\cellcolor{gray!20}\textit{Proprietary Models}} \\
    GPT-4o \cite{openai2024gpt4o} & 0.7   & 0.6   & 1.5   & 1.2   & 0.9   & 0.0   & 0.8   & 18.8  & 20.1 & 1.6 & 1.5 & 1.0 & 1.4 \\
    R-VLM \cite{park2025r-vlm} & -     & -     & -     & -     & -     & -     & -     & 66.3  & - & -     & -     & -     & - \\
    Claude Com. U. \cite{anthropic2024b-claude}  & 12.6  & 16.8  & 11.9  & 25.8  & 26.9  & 8.1   & 17.1  & 83.0  & - & -     & -     & -     & - \\
    \midrule
    \multicolumn{14}{l}{\cellcolor{gray!20}\textit{Open-source Models}} \\
    UI-TARS-72B \cite{qin2025ui-tars} & 40.8  & 39.6  & 17.2  & 45.7  & 54.8  & 30.1  & 38.1  & 88.4  & 90.3 & 31.4 & 30.5 & 14.7 & 25.2 \\
    Qwen2.5-VL-72B \cite{bai2025qwen25vl}  & 53.5  & 44.9  & 44.4  & 59.1  & 72.6  & 49.5  & 53.3  &  87.1  & 86.5 & 30.6  & 29.2 & 17.2  & 25.4  \\ 	
    JEDI-7B \cite{xie2025jedi} & -     & -     & -     & -     & -     & -     & 39.5  & -     & 91.7 & -     & -     & -     & -  \\
    SE-GUI-7B \cite{yuan2025se-gui} & 44.5  & 37.2  & 42.1  & 54.7  & 70.4  & 38.8  & 47.2  & 88.2  & 90.3 & 21.0     & 20.1     & 8.9     & 16.4  \\
    GUI-Actor-7B \cite{wu2025guiactor} & 38.1  & 41.4  & 38.3  & 50.8  & 63.0  & 38.8  & 44.6  & 89.9  & 91.0 & 34.0 & 30.4 & 8.3 & 23.8 \\
    GUI-G$^2$-7B \cite{tang2025gui-g2}  & -     & -     & -     & -     & -     & -     & 47.5  & \textbf{92.0}  & 93.3 & 35.0     & 33.0     & 11.0     & 25.9  \\
    InfiGUI-G1-7B \cite{liu2025infigui-g1} & -     & -     & -     & -     & -     & -     & 51.9 & 	-	& 93.5 & -     & -     & - & 26.1\\
    UI-Venus-7B \cite{gu2025ui-venus} & 50.2  & 42.8  & 51.0  & 57.1  & 67.8  & 37.2  & 50.8  &    -   & \textbf{94.1} & 36.1 & 32.8 & 11.9 & 26.5 \\
    \midrule
    \multicolumn{14}{l}{\cellcolor{gray!20}\textit{Ours}} \\
    GUI-ARP-7B & \textbf{59.9}  & \textbf{52.5}  & \textbf{61.7}  & \textbf{62.2}  & \textbf{77.8}  & \textbf{54.6}  & \textbf{60.8}  & 89.3  & 91.8 & \textbf{39.6} & \textbf{35.4} & \textbf{18.6} & \textbf{30.9} \\
    \bottomrule
    \end{tabular}%
    }
  \label{tab:main_results}%
\end{table*}%

\subsection{Data Curation}
For SFT training data, we employ a two-stage inference using the ARP strategy on open-source datasets \cite{chen2024dataset23, gou2024uground, li2024dataset4, chai2024dataset5}. Samples correctly answered in a single stage are labeled easy, while those that required ARP for a correct answer are designated challenging. We then leverage Qwen-VL-Max \cite{bai2025qwen25vl} to generate Chain-of-Thought for each sample, instructing it to reflect the difficulty of samples. For the GRPO training data, we use the same filtering pipeline without CoT. In total, we collect 2.4K samples for each training phase.

\section{EXPERIMENT}
\label{sec:exp}
\subsection{Experimental Setup}
We employ GUI-Actor-7B \cite{wu2025guiactor} as the backbone. For the SFT phase, all parameters are unfrozen and trained for 4 epochs. We adapt the VLM-R1 \cite{shen2025vlm-r1} framework for GRPO training, unfreeze all parameters, generate 8 responses per instruction, apply KL penalty $\beta = 0.04$, set $\alpha = 0.5$ and train for 2 epochs. During inference, we set $\tau = 0.3$, $k=20$ in ARP and instruct the model not to output its thinking process for efficiency. All experiments are conducted on 8 NVIDIA H20-96G GPUs. Following \cite{wu2025guiactor,lu2025ui-r1}, we evaluate 4 widely used GUI grounding benchmarks: ScreenSpot-v1 (SS-v1) \cite{cheng2024seeclick-ss-v1}, ScreenSpot-v2 (SS-v2) \cite{wu2024atlas-ssv2} which cover general GUI scenarios across desktop, mobile, and web interfaces, ScreenSpot-Pro (SS-Pro) \cite{li2025ss-pro}, and UI-Vision \cite{nayak2025ui-vision} with professional high-resolution software screenshots. 
\begin{table}[t]
    \centering
    \caption{Ablation study on our GUI-ARP.}
    \resizebox{\columnwidth}{!}{
    \begin{tabular}{lcccc}
    \toprule
        \textbf{Model} & \textbf{SS-Pro Avg.} & \textbf{Tool Call (\%)} & \textbf{SS-v2 Avg.} & \textbf{Tool Call (\%)}\\ 
        \midrule
        Baseline & 44.6  & 0.0  & 91.0  & 0.0  \\ 
        \cellcolor{gray!20}+ SFT & \cellcolor{gray!20}54.3  & \cellcolor{gray!20}29.7  & \cellcolor{gray!20}91.6  & \cellcolor{gray!20}11.5  \\ 
        \quad Single stage & 50.4  & 0.0  & 92.0  & 0.0  \\ 
        \cellcolor{gray!20}+ GRPO & \cellcolor{gray!20}60.8  & \cellcolor{gray!20}92.4  & \cellcolor{gray!20}91.8  & \cellcolor{gray!20}33.4 \\ 
        \quad Single stage & 50.7  & 0.0  & 92.1  & 0.0  \\ 
        \bottomrule
    \end{tabular}
    \label{tab:ablation}
    }
    \vspace{-0.3cm}
\end{table}
\subsection{Experimental Results}
We present the quantitative results in Table~\ref{tab:main_results}. The experimental results demonstrate that our method achieves state-of-the-art performance among 7B models on the challenging benchmarks SS-Pro and UI-Vision, while maintaining competitive accuracy on the SS-v1 and SS-v2 datasets. Specifically, our approach outperforms the current state-of-the-art InfiGUI-G1-7B \cite{liu2025infigui-g1} by $17.1\%$ in accuracy, and improves upon the baseline GUI-Actor \cite{wu2025guiactor} by $36.3\%$ on the SS-Pro dataset. On the UI-Vision dataset, our method shows a performance increase of $16.6\%$ compared to UI-Venus \cite{gu2025ui-venus}. 
Furthermore, our method surpasses the existing multi-stage method R‑VLM \cite{park2025r-vlm} on SS‑v1, achieving a $35\%$ higher accuracy. Notably, our 7B model also outperforms both 72B and proprietary models, despite their significantly larger parameter counts and higher training costs. This highlights the effectiveness and efficiency of our proposed GUI-ARP method.

\subsection{Ablation Study}
We perform an ablation study of the proposed method on the SS-Pro and SS-v2 benchmarks. Since ARP and ASC are designed to operate synergistically to realize adaptive region perception, we compare the baseline and our GUI‑ARP after both SFT and GRPO training, as reported in Table~\ref{tab:main_results}.
After SFT, GUI‑ARP shows a clear improvement in grounding accuracy, particularly on SS‑Pro, along with a higher tool call rate, indicating that the model develops preliminary stage controlling capability. 
With GRPO, GUI-ARP further refines its ability to assess task complexity and adaptively regulate tool invocation. On SS-Pro, our model maintains a high tool call rate to ensure robust handling of complex grounding instances. Conversely, it suppresses unnecessary tool invocation for simpler cases on SS-v2. This adaptive behavior demonstrates an effective trade-off between grounding accuracy and efficiency.
We further evaluate single‑stage localization performance. Accuracy improves consistently after two training phases. On SS-v2, single-stage version achieves even higher accuracy, as it retains more global visual information in simple structured layouts.

\section{CONCLUSION}
\label{sec:con}
We propose GUI‑ARP, a novel GUI grounding framework that leverages adaptive region perception and adaptive stage controlling strategies. Our method dynamically decides whether multi‑stage fine‑grained localization is required based on task complexity. Compared with existing multi‑stage methods, GUI‑ARP leverages internal attention distribution of the model to extract foreground regions that better align with its visual perception for further observation. We design a two-phase training pipeline combining SFT and GRPO to enhance both performance and generalization. Experimental results show that GUI‑ARP consistently outperforms existing methods with comparable parameter scales, particularly on challenging benchmarks SS-Pro and UI‑Vision.
\vfill\pagebreak

\clearpage
\begin{spacing}{0.8} 
    \bibliographystyle{IEEEbib}
    \bibliography{strings,refs}   
\end{spacing}

\end{document}